\documentclass[10pt,twocolumn,letterpaper]{article}

\usepackage{iccv}
\usepackage{times}
\usepackage{epsfig}
\usepackage{graphicx}
\usepackage{amsmath}
\usepackage{amssymb}
\usepackage{mathtools}
\usepackage{multirow}
\usepackage[inline]{enumitem}

\newcommand\iidsim{\overset{\text{i.i.d.}}{\sim}}

\usepackage[skip=3pt]{subcaption}
\newcommand{\executeiffilenewer}[3]{%
\ifnum\pdfstrcmp{\pdffilemoddate{#1}}%
{\pdffilemoddate{#2}}>0%
{\immediate\write18{#3}}\fi%
}
\newsavebox\CBox
\newcommand{\kl}[2]{D_{\rm{KL}}\left(#1\;\middle\|\;#2\right)}
\def\textBF#1{\sbox\CBox{#1}\resizebox{\wd\CBox}{\ht\CBox}{\textbf{#1}}}

\usepackage{array}
\newcolumntype{L}[1]{>{\raggedright\let\newline\\\arraybackslash\hspace{0pt}}m{#1}}
\newcolumntype{C}[1]{>{\centering\let\newline\\\arraybackslash\hspace{0pt}}m{#1}}
\newcolumntype{R}[1]{>{\raggedleft\let\newline\\\arraybackslash\hspace{0pt}}m{#1}}

\newcommand{\abs}[1]{\left\lvert#1\right\rvert}
\newcommand{\norm}[1]{\left\lVert#1\right\rVert}

\usepackage[pagebackref=true,breaklinks=true,letterpaper=true,colorlinks,bookmarks=false]{hyperref}

\iccvfinalcopy % *** Uncomment this line for the final submission

\def\iccvPaperID{10} % *** Enter the ICCV Paper ID here

\ificcvfinal\fi

\begin{document}

%%%%%%%%% TITLE
\title{Image Deconvolution with Deep Image and Kernel Priors}

\author{Zhunxuan Wang \qquad Zipei Wang \qquad Qiqi Li \qquad Hakan Bilen\\
School of Informatics, The University of Edinburgh\\
%10 Crichton St, Edinburgh EH8 9AB\\
{\tt\small \{z.wang-132, z.wang-129, q.li-50\}@sms.ed.ac.uk, hbilen@ed.ac.uk}
}

\maketitle
% Remove page # from the first page of camera-ready.
\ificcvfinal\thispagestyle{empty}\fi

%%%%%%%%% ABSTRACT
\begin{abstract}
Image deconvolution is the process of recovering convolutional degraded images, which is always a hard inverse problem because of its mathematically ill-posed property. On the success of the recently proposed deep image prior (DIP), we build an image deconvolution model with deep image and kernel priors (DIKP). DIP is a learning-free representation which uses neural net structures to express image prior information, and it showed great success in many energy-based models, e.g. denoising, super-resolution, inpainting. Instead, our DIKP model uses such priors in image deconvolution to model not only images but also kernels, combining the ideas of traditional learning-free deconvolution methods with neural nets. In this paper, we show that DIKP improve the performance of learning-free image deconvolution, and we experimentally demonstrate this on the standard benchmark of six standard test images in terms of PSNR and visual effects. %Deep image prior (DIP) is a representation which uses neural net structures to express image prior information. This work mainly focuses on the task of image deconvolution, using DIP to express image prior information and refine the domain of its objectives. It proposes new energy functions for kernel-known and blind deconvolution respectively in terms of DIP, and uses hourglass ConvNet as the DIP structures to represent sharpness or other higher level priors of natural images, as well as the prior in degradation kernels. Experiment results on 6 standard test images prove that DIP with ConvNet structure is highly capable of capturing natural image prior in image deconvolution, and that our DIP models significantly improve deconvolution performance compared with the baselines in both kernel-known and blind settings, in terms of either PSNR metric or deconvolution visual effects.
\end{abstract}

%%%%%%%%% BODY TEXT
\section{Introduction}
\label{sec:intro}
Image restoration is a long studied and challenging problem that aims to restore a degraded image to its original form \cite{basavaprasad2014study}. One way to model the processes of image degradation is convolution with translational invariance \cite{xu2014deep}
\begin{equation}
\label{equ:dg}
\mathbf{B} = \mathbf{X} \ast \mathbf{K} + \mathbf{E}
\end{equation}
where $\mathbf{X} \in \mathbb{R}^{d\times m\times n}$ is the original image, $\mathbf{K} \in \mathbb{R}^{h\times w}$ is the convolution kernel, $\mathbf{E} \in \mathbb{R}^{d\times m\times n}$ is the additive noise, $\mathbf{B} \in \mathbb{R}^{d\times m\times n}$ is the degraded image, and $d$ denotes the number of channels in the images ($1$ for greyscale images and $3$ for color images). Image deconvolution is the process of recovering the original image $\mathbf{X}$ from the observed degraded image $\mathbf{B}$, i.e. the inverse process of convolutional image degradation. This work focuses on image deconvolution in two different settings: kernel-known and kernel-unknown (a.k.a. blind deconvolution).

\textbf{Kernel-known:} The preliminary stage of image deconvolution mainly considers the case where the convolution kernel is given \cite{sezan1990survey}, i.e. recovering $\mathbf{X}$ \textit{with} knowing $\mathbf{K}$ in \autoref{equ:dg}. This problem is ill-posed, because simply applying the inverse of the convolution operation on degraded image $\mathbf{B}$ with kernel $\mathbf{K}$, i.e. $\mathbf{B} \ast^{-1} \mathbf{K}$, gives an inverted noise term $\mathbf{E} \ast^{-1} \mathbf{K}$, which dominates the solution \cite{hansen2006deblurring}.

\textbf{Blind deconvolution:} In reality, we can hardly obtain the detailed kernel information and the deconvolution problem is formulated in a \textit{blind} setting \cite{kundur1996blind}. More concisely, blind deconvolution is to recover $\mathbf{X}$ \textit{without} knowing $\mathbf{K}$. This task is much more challenging than it is under non-blind settings, because the observed information becomes less and the domains of the variables become larger \cite{chan1998total}.

In image deconvolution, \textit{prior} information on unknown images and kernels (in blind settings) can significantly improve the deconvolved results. A traditional representation for such prior information is handcrafted regularizers in image energy minimization \cite{fornasier2010convergent}, e.g. total variation (TV) regularization for image sharpness \cite{chan1998total} and $L^1$ regularization for kernel sparsity \cite{shan2008high, wang2017iterative}. However, prior representations like the above-mentioned regularizers have limited ability of expressiveness \cite{majumdar2009classification}.
Therefore, this work aims to find better prior representations of images and kernels to improve deconvolution performances.

Deep neural architecture has a strong capability to accommodate and express information because of its intricate and flexible structure \cite{szegedy2013intriguing}. Compared to other image prior representations with limited structures (e.g. regularizers), neural nets with such powerful expressiveness seem more capable of capturing higher-level prior of natural images and degradation kernels. Deep image prior (DIP) \cite{ulyanov2018deep} is a neural-based image prior representation which achieved good performance in various image restoration problems. The main idea of DIP is to substitute image variable in an energy function by the output of a deep convolutional neural net (ConvNet) with random noise inputs, so that the image prior can be captured by the \textit{hyperparameter} of the ConvNet, and the output image is determined by the \textit{parameter} of the ConvNet. One point to emphasize here is that priors expressed by both handcrafted regularizers and DIP are embodied in their own formulations or structures, which does not require large datasets for training. In the existing applications (incl. denoising, inpainting, etc.) of DIP, the degradation processes are considered as known. In this paper, we are the first to show that deep priors perform well in image deconvolution. Furthermore, we show that ConvNets can be utilized as a source of prior knowledge not only for natural images but also for degradation kernels (named as deep kernel prior, DKP), bridging the gap between traditional methods and deep neural nets. Through experiments we demonstrate that our deep image and kernel priors (DIKP) result in a significant improvement over traditional learning-free regularization-based priors in image deconvolution\footnote{We do not show any results from supervised deep network techniques because our method is unsupervised and our objective is to prove that our deep priors are better than handcrafted priors in image deconvolution.}.
%In this paper, we show that \begin{enumerate*} [label=\itshape\alph*\upshape)] \item DIP also performs well in image deconvolution, b) deep prior is also capable of modeling degradation kernels in blind deconvolution (named as DKP). \end{enumerate*} Compared to handcrafted regularizers, deep image and kernel priors (DIKP) express prior information by their network structures, and explore solutions in the \textit{ConvNets' parameter spaces} instead of the \textit{image/kernel spaces}. One point to emphasize here is that the priors expressed by both handcrafted regularizers and DIKP are embodied in their own formulations or structures, which does not require large datasets for training.%For kernel variables in blind deconvolution, we also adopt ConvNets as kernel priors. More specifically, DIP image deconvolution \begin{enumerate*} [label=\itshape\alph*\upshape)] \item finds a hyperparameter setting for the ConvNet which is capable of expressing image prior implicitly, then \item searches for the optimal parameter on the ConvNet that minimizes the energy. \end{enumerate*}

%To summarize, we propose the general energy function for DIKP image deconvolution (see \autoref{sec:method}). Our model achieves significantly better performance than the learning-free baselines we set in different degradation kernels (see \autoref{sec:expts}), especially in motion blur (see \autoref{fig:comp_cmm}).

%------------------------------------------------------------------------

\section{Related work}
\label{sec:relwork}
%This section should review published work which can help to give a better understanding of your work -- related approaches, other work on the same data, ideas for future work. The aim is to try to place what you have done in a wider context.
The earliest traditional methods of image deconvolution include Richardson-Lucy (RL) method \cite{richardson1972bayesian} and Weiner Filtering \cite{wiener1949extrapolation}. Due to their simplicity and efficiency, these two methods are still widely used today, but they may be subject to ringing artifacts \cite{proakis2001digital}. To solve this, many refinements based on handcrafted regularization priors came out. \cite{dey2006richardson} adopted TV regularizer as prior in kernel-known deconvolution. \cite{yuan2008progressive} proposed a progressive multi-scale optimization method based on RL method, with edge-preserving regularization as the image prior. For degradation kernels, early methods \cite{reeves1992blur} only dealt with their simple parametric forms. Later then, natural image statistics were used to estimate kernels \cite{fergus2006removing, levin2007blind}. After that, \cite{shan2008high, wang2017iterative} adopted $L^1$ regularizer as kernel prior in blind deconvolution. However, handcrafted priors mentioned above have relatively simple structures, so their expressiveness is rather limited \cite{majumdar2009classification}.

This work is inspired by traditional image deconvolution methods by handcrafted priors \cite{rudin1992nonlinear, wang2017iterative}, but trying to use deep image priors instead of handcrafted priors. It uses ConvNet to express the prior information of both natural images and degradation kernels, putting kernel-known and blind deconvolution under the same model. Besides, as discussed in \cite{ulyanov2018deep}, its ConvNet-based image prior representation links two sets of popular deconvolution methods: learning-based approaches by ConvNet \cite{xu2014deep, zhang2017learning, MaoSY16} and learning-free approaches by handcrafted prior \cite{shan2008high}.

%------------------------------------------------------------------------

\section{Data set and evaluation metrics}
\label{sec:data}
\begin{figure}[tb]
\centering
\includegraphics[width=.95\columnwidth]{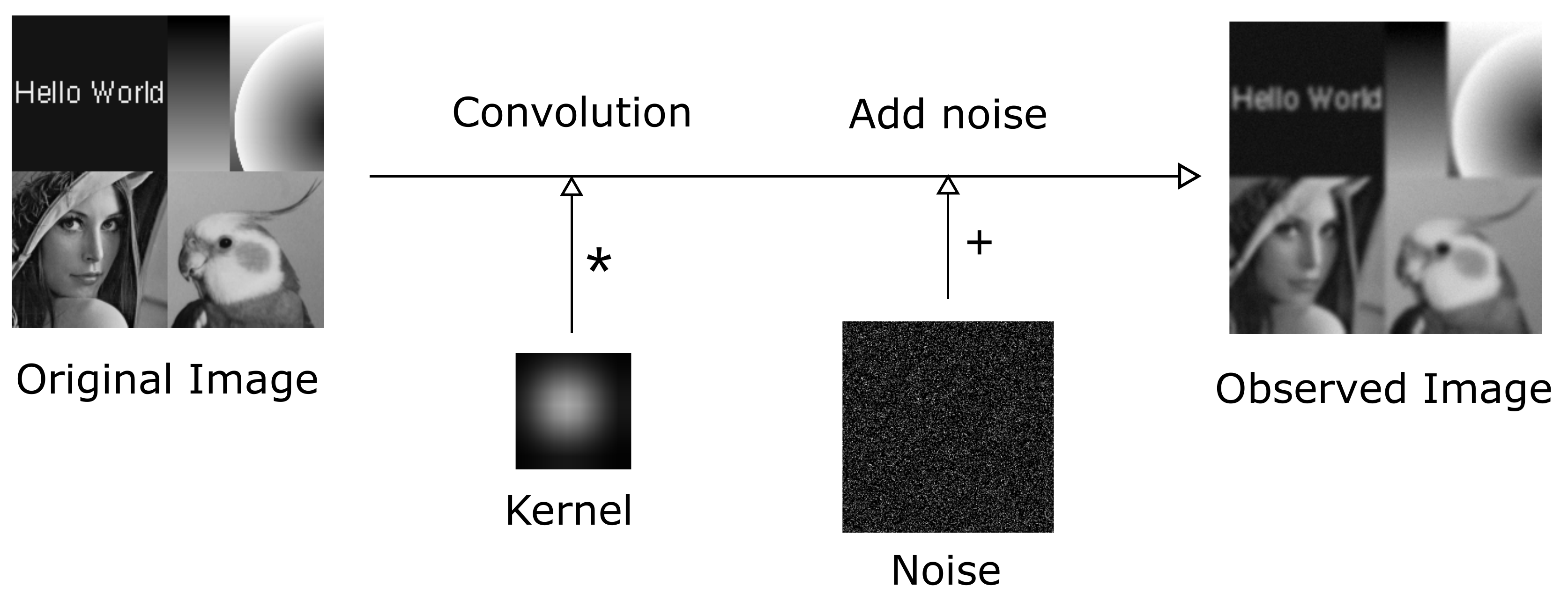}
\caption{\textbf{The generation processes of observed images}. For each process, we first convolve the original image by a given kernel, then add a noise term to the convolved image.}\label{fig:data_gen}
\end{figure}

\begin{figure}[tb]
\centering
\minipage{0.16\columnwidth}
  \includegraphics[width=\linewidth]{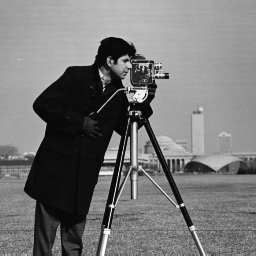}
  \subcaption*{\scriptsize{$256\times 256$}}
\endminipage\hfill
\minipage{0.16\columnwidth}
  \includegraphics[width=\linewidth]{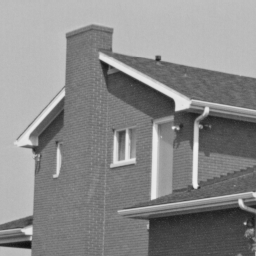}
  \subcaption*{\scriptsize{$256\times 256$}}
\endminipage\hfill
\minipage{0.16\columnwidth}%
  \includegraphics[width=\linewidth]{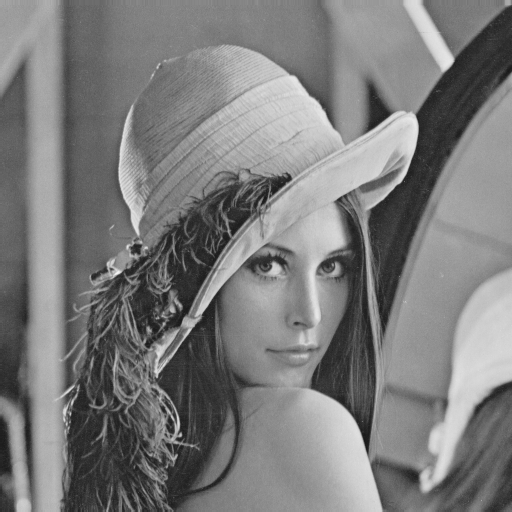}
  \subcaption*{\scriptsize{$512\times 512$}}
\endminipage\hfill
\minipage{0.16\columnwidth}
  \includegraphics[width=\linewidth]{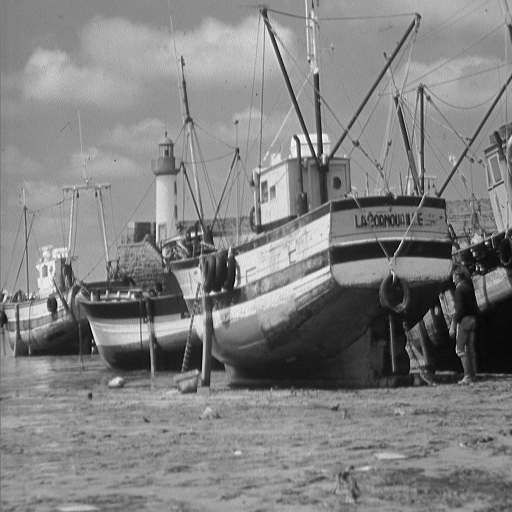}
  \subcaption*{\scriptsize{$512\times 512$}}
\endminipage\hfill
\minipage{0.16\columnwidth}
  \includegraphics[width=\linewidth]{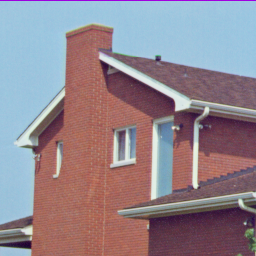}
  \subcaption*{\scriptsize{$256\times 256$}}
\endminipage\hfill
\minipage{0.16\columnwidth}%
  \includegraphics[width=\linewidth]{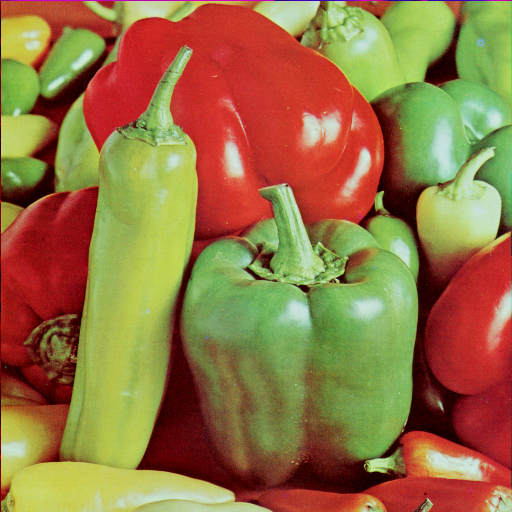}
  \subcaption*{\scriptsize{$512\times 512$}}
\endminipage
\caption{\textbf{Standard test image data set experimented in our work} (zoomed out), containing $4$ greyscale and $2$ color images, named \texttt{cameraman} (abbr. \texttt{C.man}), \texttt{house}, \texttt{Lena}, \texttt{boat}, \texttt{house.c}, \texttt{peppers} respectively from left to right. The original resolutions are marked below them.}\label{fig:std_data}
\end{figure}
As discussed in \autoref{sec:intro}, capturing image prior by either regularization or deep neural net structures is learning-free. Therefore, data set explored in this work is only used for testing. Experiments and performance evaluation are conducted on a data set with $6$ standard test images shown in \autoref{fig:std_data}. Those images, along with their preprocessing and evaluation mentioned in the following, are in line with standard practice and widely used in denoising \cite{dabov2007video}, TV deblurring \cite{beck2009fast}, etc., which guarantees the reliability of our results.

\subsection{Observed data generation and kernels}
To preprocess the image data and obtain degraded observations, we use the degradation model formulated as \autoref{equ:dg} to transfer the original standard test image $\mathbf{X}_{\rm{std}}$ to the observed image $\mathbf{B}$, illustrated by the diagram in \autoref{fig:data_gen}. The noise matrix $\mathbf{E}$ is i.i.d. Gaussian with respect to each entry, and the noise strength (i.e. standard deviation) $\sigma$ is fixed at $0.01$ to reduce experimental variables. To explore different kinds of degradation models, three common kernels for different kinds of degradation, Gaussian kernel \cite{hummel1987deblurring}, defocus \cite{hansen2006deblurring} and motion blur \cite{yitzhaky1997identification} are used to generate the data set.

\textbf{Gaussian:} The kernel for degradation caused by atmospheric turbulence can be described as a two-dimensional Gaussian function \cite{jain1989fundamentals, roggemann2018imaging}, and the entries of the unscaled kernel are given by the formula \cite{hansen2006deblurring}
$$K_{i,j} = \exp\left[-\frac12\left(\frac{i - c_1}{s_1}\right)^2 - \frac12\left(\frac{j - c_2}{s_2}\right)^2\right]$$
where $\left(c_1, c_2\right)$ is the center of $\mathbf{K}$, and $\left(s_1, s_2\right)$ determines the width of the kernel (i.e. standard deviation of the Gaussian). In this work, $s_1$ and $s_2$ are set to $s_1 = s_2 = 2.0$.

\textbf{Defocus:} Out-of-focus is another issue in optical imaging. Knowledge of the physical process that causes out-of-focus provides an
explicit formulation of the kernel \cite{hansen2006deblurring}
\begin{equation*}
K_{i,j} = \begin{cases}
1/(\pi r^2) & \text{if $\left(i - c_1\right) ^ 2 + \left(j - c_2\right) ^ 2 \leq r^2$,}\\
0 & \text{otherwise.}
\end{cases}
\end{equation*}
where $r$ denotes the radius of the kernel, which is set to $r = \lfloor\min\left(h / 2, w / 2\right)\rfloor$ in this work.

\begin{figure}[tb]
\minipage{0.25\columnwidth}
  \includegraphics[width=\linewidth]{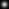}
  \subcaption{Gaussian}
\endminipage\hfill
\minipage{0.25\columnwidth}
  \includegraphics[width=\linewidth]{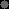}
  \subcaption{Defocus}
\endminipage\hfill
\minipage{0.25\columnwidth}%
  \includegraphics[width=\linewidth]{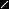}
  \subcaption{Motion blur}\label{fig:kernel_motion}
\endminipage
\caption{Visualization examples of the $3$ kernels.}\label{fig:kernels}
\end{figure}

\textbf{Motion blur:} This happens if an image being recorded changes in a single exposure when taking a photograph. For example, when taking a picture, moving objects being taken at high speed or lens shake will blur the picture. In noiseless case, the convolution processes of motion blur with amplitude $u$ and shifting angle $\alpha$ are given by the formula \cite{kalalembang2009dct}
$$B_{i,j} = \frac1{2u+1} \sum_{k=-u}^u X_{i + k\cos\alpha, j + k\sin\alpha}$$
in which the shape of the kernel is a line segment as \autoref{fig:kernel_motion} shows. In this work, the blur amplitude and shifting angle are set as $u = \sqrt{2}\cdot\lfloor\min\left(h/2,w/2\right)\rfloor$ and $\alpha = 3\pi/4$.

All the kernels adopted in data generation processes and experiments in this work are in shape $9\times 9$ (i.e. $h = w = 9$) with center $\left(4, 4\right)$ (i.e. $c_1 = c_2 = 4$), and scaled such that elements in each kernel sum to $1$ \cite{hansen2006deblurring}. \autoref{fig:kernels} gives visualization examples of the $3$ different kernels adopted with given settings mentioned above.
\subsection{Evaluation metrics}
We use the \textit{Mean Square Error} (\textit{MSE}) between the degraded image variable $\mathbf{B}_{\rm{var}} = \mathbf{X} \ast \mathbf{K}$ and the observation
$$\operatorname{MSE}\left(\mathbf{B}_{\rm{var}}, \mathbf{B}\right) = \frac{1}{d\cdot m\cdot n}\norm{\mathbf{B}_{\rm{var}} - \mathbf{B}}_2^2$$
to measure the energy function \cite{ulyanov2018deep} and to track parameter iterations in the first experiment (see \autoref{subsec:cvg}). Using this metric, to minimize the energy is to find the image $\mathbf{X}$ that, when degraded, is the same as the observation $\mathbf{B}$.

To measure image deconvolution quantitatively, we use the \textit{Peak Signal to Noise Ratio} (\textit{PSNR}) (in dB) \cite{huynh2008scope} between the image variable $\mathbf{X}$ and the standard test image $\mathbf{X}_{\rm{std}}$
$$\operatorname{PSNR}\left(\mathbf{X}, \mathbf{X}_{\rm{std}}\right) = 10\log_{10}\left[\frac{R^2}{\operatorname{MSE}\left(\mathbf{X}, \mathbf{X}_{\rm{std}}\right)}\right]$$
where $R$ is the maximum possible pixel value of the image, e.g. $R = 1$ if images in double-precision floating-point data type, $R = 255$ if in $8$-bit data type. In this work, we use double-precision floating-point data type, i.e. $R = 1$.

In \autoref{subsec:exp_grad}, we compare the gradient distributions among output images and standard test images. To measure the similarity between a gradient frequency distribution $\Pr\left(\cdot\right)$ and one by standard test images $\Pr_{\rm{std}}\left(\cdot\right)$, we use the \textit{Kullback-Leibler} (\textit{KL}) divergence \cite{kullback1997information}
$$\kl{\Pr}{\Pr\nolimits_{\rm{std}}} = -\sum_{b\,\in\,\mathcal{B}}\Pr\left(b\right)\log\frac{\Pr\nolimits_{\rm{std}}\left(b\right)}{\Pr\left(b\right)}$$
where $b$ denotes a bin corresponding to a range of gradient values, $\mathcal{B}$ is the whole bin set covering all possible gradient values. From the definition, the similarity between two distributions and their \textit{KL} divergence are negatively correlated.

%------------------------------------------------------------------------

\section{Methodology}
\label{sec:method}
According to \autoref{sec:intro}, both regularization-based prior and deep image prior are embedded in energy minimization models, which, in general, are formulated as \cite{fornasier2010convergent}
\begin{equation}
\label{equ:engy_trad}
\min_{\mathbf{X}} E\left(\mathbf{X}; \, \mathbf{B}\right) + R\left(\mathbf{X}\right)
\end{equation}
where $E\left(\mathbf{X}; \, \mathbf{B}\right)$ indicates the energy term associated with the data, and $R\left(\mathbf{X}\right)$ is the prior term. A general explanation of the energy term is the \textit{numerical difference} between the given image data and the image variable processed by given degradation. For image deconvolution, the degradation operator is convolution, therefore the energy is designed as $E\left(\mathbf{X};\, \mathbf{B}\right) = \operatorname{MSE}\left(\mathbf{X} \ast \mathbf{K}, \mathbf{B}\right)$. The energy term $E\left(\mathbf{X}; \, \mathbf{B}\right)$ can also be designed for other tasks in image restoration, such as inpainting \cite{shen2003euler}, super-resolution \cite{gerchberg1974super} and image denoising \cite{rudin1992nonlinear}. Methods adopted in this work are all based on the deconvolution energy model and its mutants.
%In this section, we will introduce technical methodologies in detail.
%For our research problem, the main method of deep image prior  and image deconvolution will be introduced. Additionally, how to apply the main idea of deep image prior on the image deconvolution will be presented. Finally,we will introduce evaluation methods for image deconvolution task.
\subsection{Baseline models with regularization prior}
\label{subsec:baseline}
The gradient magnitude of a two-dimensional function $x\left(s, t\right)$ is defined and formulated as the following \cite{gonzalez1977digital}
$$\abs{\nabla x}\left(s, t\right) = \norm{\nabla x\left(s, t\right)}_2 = \sqrt{\left(\frac{\partial x}{\partial s}\right)^2 + \left(\frac{\partial x}{\partial t}\right)^2}\text{,}$$
the discrete formulation of which for an image $\mathbf{X}$ is given by the following matrix
$$\abs{\nabla\mathbf{X}} = \sqrt{\left(\mathbf{X}\mathbf{D}_{1, n}^\top\right)^2 + \left(\mathbf{D}_{1, m}\mathbf{X}\right)^2}$$
where the square and the square root calculations are entry-wise, and $\mathbf{D}_{1, n}$ is the discrete partial derivative operator (see \cite[Chap.~7]{hansen2006deblurring} and \cite[Sec.~2]{chambolle2004algorithm} for its formal definition and its specified usage in this paper, respectively).
%$$\mathbf{D}_{1, n} = \begin{bmatrix}
%1 & -1 & & \\
% & \ddots & \ddots & \\
% & & 1 & -1 \\
% & & & 0
%\end{bmatrix}_{n \times n}\text{.}$$

In image processing, discrete gradient magnitudes are proven to be a strong prior to natural images \cite{shan2008high, hansen2006deblurring}. The sum of such magnitudes in a single image is a regularization representation of the image prior, i.e. total variation norm
$$\norm{\mathbf{X}}_{\rm{TV}} \coloneqq \sum\limits_{i = 1}^m\sum\limits_{j = 1}^n\sqrt{\left(\mathbf{X}\mathbf{D}_{1, n}^\top\right)_{i,j}^2 + \left(\mathbf{D}_{1, m}\mathbf{X}\right)_{i,j}^2}\text{.}$$
The efficiency of TV norm has been proven for recovering blocky images \cite{dobson1996recovery} and images with sharp edges \cite{chan2000high}.

It is also known that $L^1$ norm is capable of expressing the sparsity of matrices \cite{friedman2001elements}, defined as
$$\norm{\mathbf{X}}_1 = \sum_{i=1}^m\sum_{j=1}^n\abs{X_{i, j}}\text{.}$$
In most instances, degradation convolution kernels are sparse \cite{shan2008high}. Thus $L^1$ sparsity regularization is a strong prior to convolution kernels in blind settings.

The baseline models in this work are energy minimization with TV and $L^1$ regularization priors, of which the details in the two main settings are as follows.

\textbf{Kernel-known:} The baseline model with $\mathbf{K}$ known is formulated as the following energy minimization model with TV regularization prior
\begin{equation}
\min_{\mathbf{X}} \operatorname{MSE}\left(\mathbf{X} \ast \mathbf{K}, \mathbf{B}\right) + \alpha \norm{\mathbf{X}}_{\rm{TV}}
\end{equation}
where $\alpha$ is the TV regularization parameter. To solve the TV regularization system efficiently, we adopt a fast gradient-based algorithm named \textit{MFISTA} \cite{beck2009fast} , which has performed remarkable time-efficiency and convergence property in TV regularization.

\textbf{Blind deconvolution:} The baseline system in blind setting introduces a new sparsity prior compared to the non-blind baseline above, which is formulated as
\begin{equation}
\min_{\mathbf{X}, \mathbf{K}} \operatorname{MSE}\left(\mathbf{X} \ast \mathbf{K}, \mathbf{B}\right) + \alpha \norm{\mathbf{X}}_{\rm{TV}} + \beta \norm{\mathbf{K}}_1
\end{equation}
where $\beta$ is the $L^1$ regularization parameter. This TV-$L^1$ double-prior system can be solved using \textit{TNIP-MFISTA} algorithm proposed in \cite{wang2017iterative}. To optimize both the image and the kernel, this algorithm adopts \textit{fix-update} iterations between \textit{MFISTA} and an $L^1$ regularization algorithm named \textit{Truncated Newton Interior Point method} (\textit{TNIP}) \cite{kim2007efficient}.
\subsection{Deconvolution with DIKP}
\begin{figure}
    \centering
    \includegraphics[width=\columnwidth]{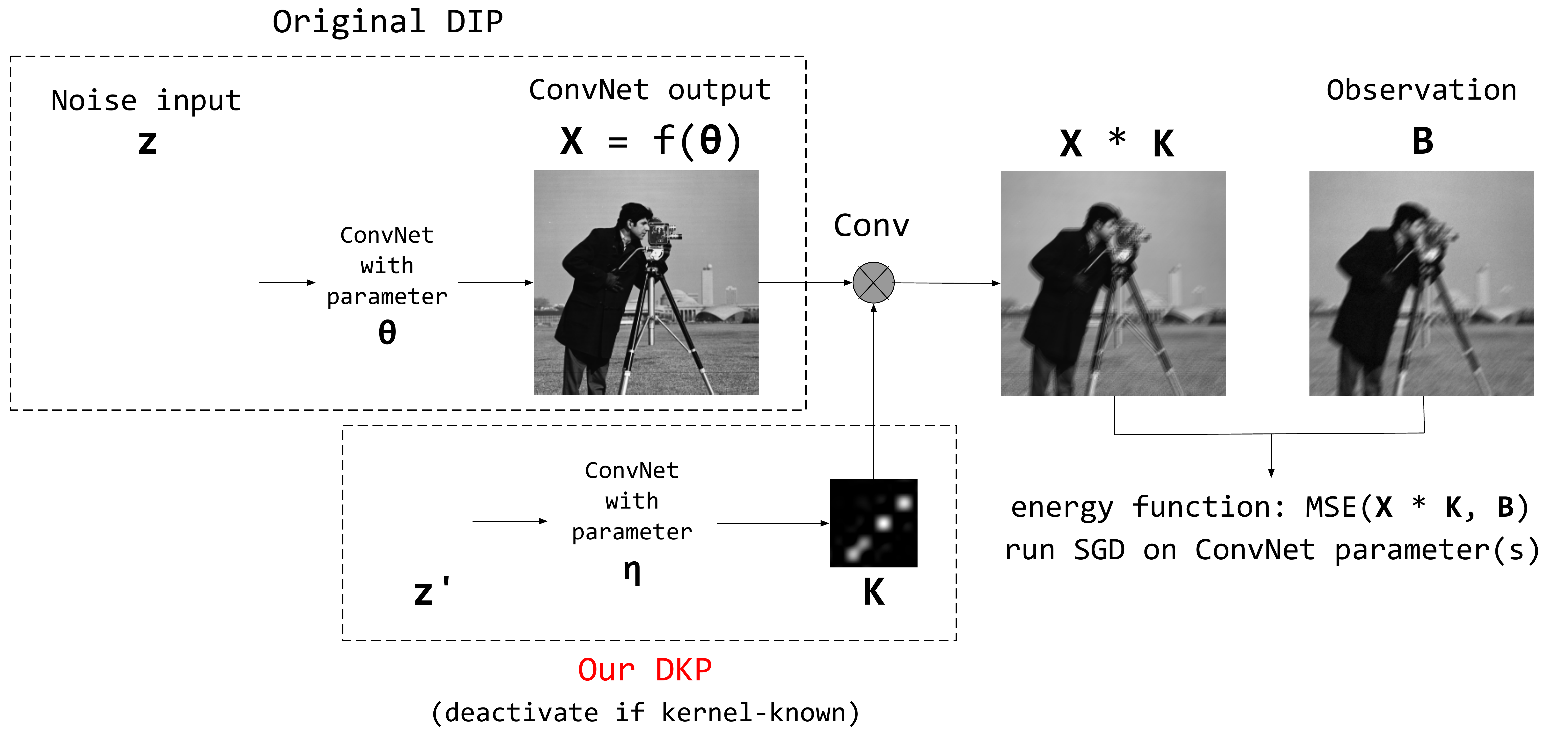}
    \caption{\textbf{The overall pipeline of our DIKP deconvolution model}, corresponding to \autoref{equ:dip_nb} and \autoref{equ:dip_blind}.}
    \label{fig:dip_pipeline}
\end{figure}
DIKP aim to capture the priors of images/kernels by the structures of generative deep neural nets. Taking image variable $\mathbf{X}$ as an example, it re-parameterises the image $\mathbf{X}$ as the neural net output $\mathbf{X} = f\left(\mathbf{z};\, \boldsymbol{\theta}\right)$, defined as the following surjection
$$f^\ast: \operatorname{supp} p \times \Theta \xmapsto[]{\text{ConvNet}} \mathcal{X},\, \left(\mathbf{z}, \boldsymbol{\theta}\right) \rightarrow \mathbf{X}$$
where $\operatorname{supp} p$ denotes the support\footnote{$\operatorname{supp} p = \left\{\mathbf{z} \in \Omega \mid p\left(\mathbf{z}\right) \neq 0\right\}$ \cite{royden1988real}, where $\Omega$ is the sample space of noise vector $\mathbf{z}$.} of the input noise probability density function $p$, $\Theta$ denotes the weight space determined by the network structure, and $\mathcal{X}$ is the solution space of $\mathbf{X}$, containing the prior information. The neural net $f^\ast$ maps the random noise network input $\mathbf{z}$ and the network weights $\boldsymbol{\theta}$ to the output $\mathbf{X}$. Ideally, by adjusting the network structure to its optimum, the solution space $\mathcal{X}$ only contains images on desired prior information.

From the perspective of mechanics, the desired prior is expressed by the network structure, and the weights $\boldsymbol{\theta}$ explores solutions on the prior. The random input noise $\mathbf{z}$ is a high-dimensional Gaussian. The main reason to take a random noise as the network input is to increase the robustness \cite{morales2007adding} to overcome degeneracy issues. On the other hand, high-dimensional Gaussian vectors are essentially concentrated uniformly in a sphere \cite{johnstone2006high}. Therefore the input space $\operatorname{supp} p$ can be approximated as a single point, and the surjection can be rewritten with the input space eliminated
$$f: \Theta \mapsto \mathcal{X},\, \boldsymbol{\theta} \rightarrow \mathbf{X}$$
%From the work of \cite{ulyanov2018deep}, we could know that the goal of image restoration task is to get original X from a corrupted image $X_0$. Such tasks are regarded as optimization tasks:
which maps only a selection of parameters $\boldsymbol{\theta}$ on the network, to an output image $\mathbf{X}$. In the rest of the report, $f\left(\boldsymbol{\theta}\right)$ denotes output image by deep image prior $f$ with weight $\boldsymbol{\theta}$.
\subsubsection{Energy functions of DIKP deconvolution}
Traditional energy minimization (formulated as \autoref{equ:engy_trad}) for image deconvolution explores the whole image space as the domain. By re-parameterising the image term $\mathbf{X}$ into the neural net output $f\left(\boldsymbol{\theta}\right)$, the solution space contains the prior information expressed by the structure of $f$, instead of the prior term $R\left(\mathbf{X}\right)$. Thereby with deep image prior the general energy model by \autoref{equ:engy_trad} turns into
\begin{equation}
\label{equ:engy_dip}
\min_{\boldsymbol{\theta}} E(f\left(\boldsymbol{\theta}\right);\, \mathbf{B})\text{.}
\end{equation}
By optimizing network weights $\boldsymbol{\theta}$ on a ideal structure, an image is optimized conditioned on the desired prior.

\textbf{Kernel-known} image deconvolution objective with deep image prior is derived directly from \autoref{equ:engy_dip}, by applying the deconvolution energy function
\begin{equation}
\label{equ:dip_nb}
\min_{\boldsymbol{\theta}} \operatorname{MSE}\left(f\left(\boldsymbol{\theta}\right)\ast\mathbf{K},\mathbf{B}\right)
\end{equation}
where $\mathbf{K}$ is the observed kernel. The minimizer $\boldsymbol{\theta}^\ast$ is obtained by \textit{Adam} optimizer \cite{kingma2014adam} with \textit{random initialization}.

\textbf{Blind deconvolution:} In blind settings, the convolution kernel $\mathbf{K}$ is assumed to be unobservable. Thereby the kernel is parameterised by another deep neural net structure $g\left(\boldsymbol{\eta}\right)$ containing prior information regarding degradation kernels. After parameterisation on kernel matrix in \autoref{equ:dip_nb}, the blind deconvolution objective with deep image prior is formulated as the following system
\begin{equation}
\label{equ:dip_blind}
\min_{\boldsymbol{\theta},\, \boldsymbol{\eta}} \operatorname{MSE}\left(f\left(\boldsymbol{\theta}\right)\ast g\left(\boldsymbol{\eta}\right),\mathbf{B}\right)
\end{equation}
where $f$ and $g$ have different ConvNet structures since the prior information of natural images and kernels are apparently different. To obtain the minimizers $\boldsymbol{\theta}^\ast$ and $\boldsymbol{\eta}^\ast$, we use \textit{Adam} to update the two variables \textit{simultaneously}. % Thus the main challenge is to find efficient structures on both $f$ and $g$.

\autoref{fig:dip_pipeline} gives a diagram summarizing our DIKP deconvolution model. Hyperparameter settings for both $f$ and $g$ are explained in detail in \autoref{sec:expts}.

%------------------------------------------------------------------------

\section{Experiments}
\label{sec:expts}
To explore to what extent deep priors can capture prior knowledge of natural images in deconvolution models, we \begin{enumerate*} [label=\itshape\alph*\upshape)]
\item compare the energy convergence property during DIKP deconvolution optimization between natural images and noise images; \item compare the gradient distributions among standard test images and images from both baseline model and DIKP.
\end{enumerate*}
This part of experiments aims to evaluate DIKP's expressiveness on natural images, therefore it is only conducted in \textit{kernel-known} setting, i.e. DKP is deactivated.

The second part of our experiment aims to find out whether our proposed DIKP deconvolution models improve the performance of image deconvolution in both \textit{kernel-known} and \textit{blind} settings, compared with the baselines. In our results, \textit{PSNR} comparison is conducted for quantitative analysis on deconvolution performance, and qualitative analysis is based on the presented images.

\subsection{Experiment Setup}
\begin{figure}[tb]
    \centering
    \includegraphics[width=0.66\columnwidth]{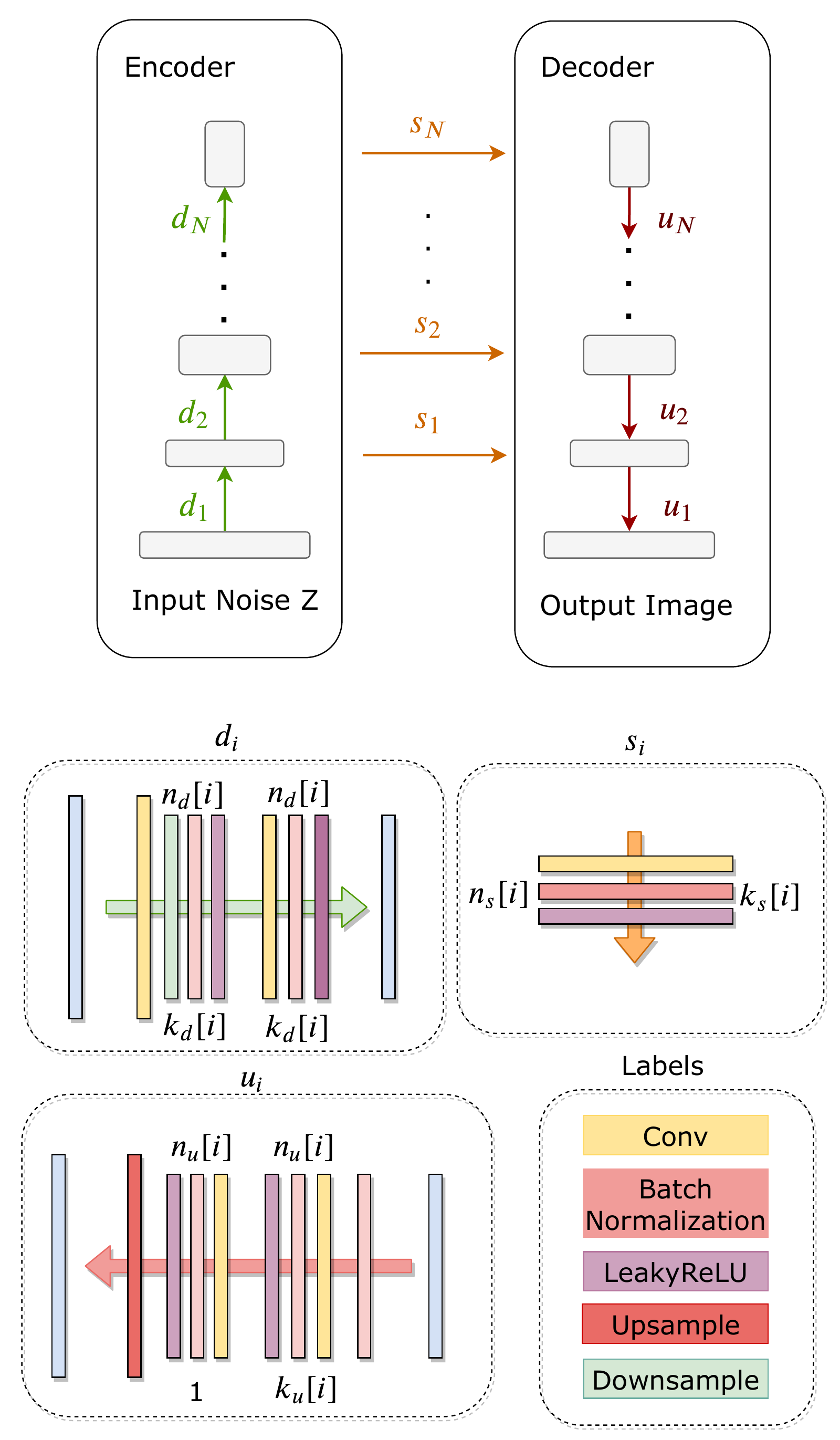}
    \caption{The \textit{hourglass} architecture as our DIKP structures. The upper half is the high-level encoder-decoder network with skip connections \cite{MaoSY16}. The detailed structures inside each downsample connection $d_i$, upsample connection $u_i$ and skip connection $s_i$ are shown below the high-level structure, where $n_u[i]$, $n_d[i]$, $n_s[i]$ denote the numbers of filters in their respective connections at depth $i$, and $k_u[i]$, $k_d[i]$, $k_s[i]$ are the corresponding kernel sizes.}
    \label{fig:hourglass}
\end{figure}
\begin{figure*}[tb]
\centering
\includegraphics[width=0.88\textwidth]{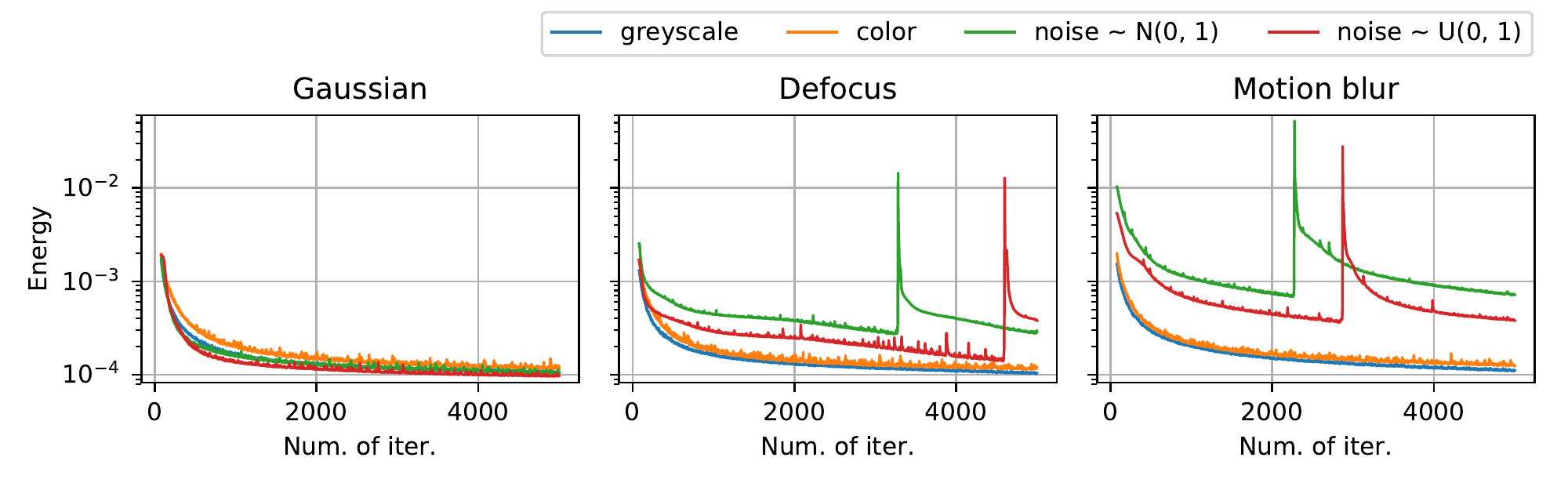}
\caption{Optimization curves of different types of images/noise for kernel-known DIKP deconvolution.}\label{fig:mse_curve}
\end{figure*}
\begin{figure}[tb]
\centering
\includegraphics[width=0.75\columnwidth]{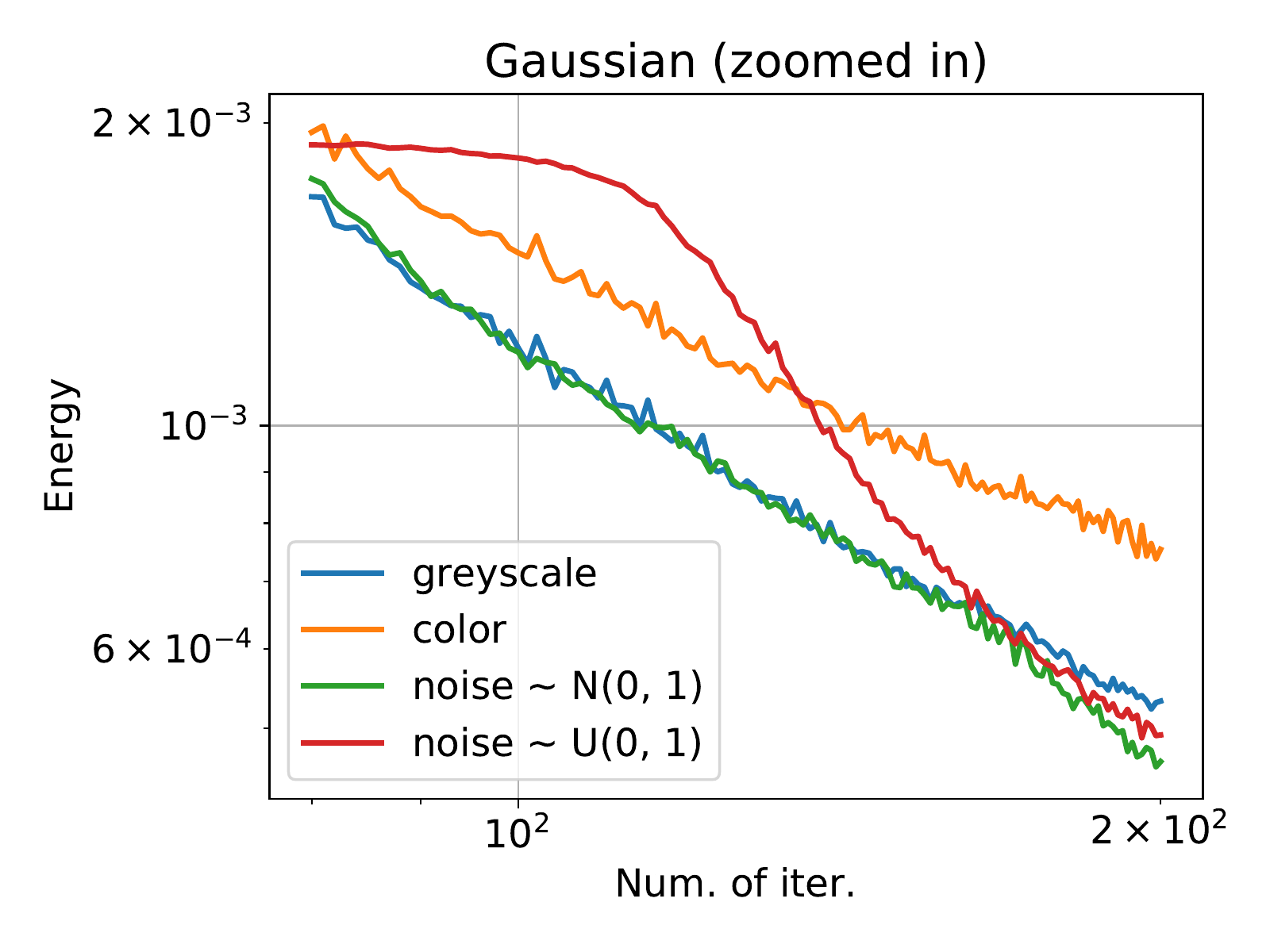}
\caption{Zoomed-in optimization curve of Gaussian case originally shown in \autoref{fig:mse_curve} (from iteration $80$ to $200$).}\label{fig:mse_curve_gauss}
\end{figure}

\textbf{Convolution}: Convolution processes in this paper, including data generation and energy calculations, are subject to \textit{reflexive} boundary condition \cite{hansen2006deblurring}. Specifically, for color images, all channels share the \textit{same} kernel \cite{hansen2006deblurring}.

\textbf{Baseline}: In kernel-known setting, the TV regularization parameter $\alpha$ is set to $2 \times 10^{-2}$, within a reasonable range for image deconvolution according to \cite{beck2009fast}. In blind setting, the regularization parameters are set to $\alpha = 2 \times 10^{-3}$ and $\beta = 5$ as the same in \cite{wang2017iterative}, among the experiments of which such setting achieved the best results.

\textbf{ConvNet architecture as DIKP}: As suggested for super-resolution setting in \cite{ulyanov2018deep}, we use \textit{hourglass} architecture (shown in \autoref{fig:hourglass}) as the main body of DIKP, whose hyperparameter settings are shown as follows

\noindent
\fbox{%
\begin{minipage}[t]{0.45\columnwidth}
For images:\\
$\mathbf{z} \iidsim \mathcal{U}\left(0, 0.1\right)$;\\
$n_u = n_d = [128] \times 5$;\\
$k_u = k_d = [3,3,3,3,3]$;\\
$n_s = [4,4,4,4,4]$;\\
$k_s = [1,1,1,1,1]$;\\
upsample stride size $2$;\\
Sigmoid to output.
\end{minipage}}%
\hfill%
\fbox{%
\begin{minipage}[t]{0.45\columnwidth}
For kernels (if blind):\\
$\mathbf{z} \iidsim \mathcal{U}\left(0, 0.1\right)$;\\
$n_u = n_d = [128] \times 5$;\\
$k_u = k_d = [3,3,3,3,3]$;\\
$n_s = [4,4,4,4,4]$;\\
$k_s = [1,1,1,1,1]$;\\
upsample stride size $1$;\\
Softmax to output.
\end{minipage}
}

\noindent
We put Sigmoid and Softmax on ConvNet outputs for images and kernels respectively, because image pixels range from $\left[0, 1\right]$ and kernel pixels sum to $1$. The reason for setting upsample stride size to $1$ for kernel generation is to prevent degeneration due to their small size ($9\times 9$). It is worth mentioning that we apply add-noise regularization to the neural network, i.e. we disturb the noise input $\mathbf{z}$ with an additive Gaussian $\mathbf{z} \gets \mathbf{z} + \Delta\,\mathbf{z}$ at the beginning of each iteration. This technique aims to increase model robustness to perturbation \cite{morales2007adding}. Although this regularization has a negative impact on the optimization process, we find that the network can still converge the energy to $0$ with a sufficient number of iterations and improve deconvolution performance.

\subsection{Bias in convergence}\label{subsec:cvg}
Even though the complex structure of the neural network in a DIKP model allows the solution space to have a variety of features regarding natural images, it is still possible for the DIKP model to express interference information other than natural images \cite{szegedy2013intriguing}, e.g. noise. Therefore, we introduce noise into our experiments, using our DIKP kernel-known model on natural images (incl. greyscale and color images) and noise respectively. By comparing the convergence property of the energy functions on the two in the optimization process, we can know whether our model can block such interference information in its solution space.

In our control experiment, we decide to use Gaussian white noise and uniform noise, generated from Gaussian $\mathcal{N}\left(0, 1\right)$ and uniform $\mathcal{U}\left(0, 1\right)$. \autoref{fig:mse_curve} shows the optimization curves of energy values with respect to iterations in DIKP kernel-known deconvolution, where each plot corresponds to each degradation kernel. In spite of the Gaussian kernel, energy value convergence shows obvious differences between natural images and noise in DIKP deconvolution with defocus and motion blur kernels. More specifically, we observe that curves by the noise are clearly above those by natural images, and sudden leaps take place for energy values by noise in both plots. We speculate, the cause of this observation is that, the ConvNet structures in DIKP are unstable to parameter fluctuations for noise generation, which also explains how DIKP deconvolution blocks noise information. For the Gaussian, although in \autoref{fig:mse_curve} we cannot see a wild difference between noise and natural images, in \autoref{fig:mse_curve_gauss} we can still observe that the energy value by the uniform noise converges slower than that by natural images in early iterations, which also indicates that DIKP model blocks uniform noise in Gaussian degraded deconvolution.

The DIKP deconvolution in the control experiments with noise indeed shows biases to natural images from the perspective of energy function convergence, which means in most cases, DIKP are capable of blocking interference and irrelevant information in image deconvolution.

\subsection{Image gradient distributions}\label{subsec:exp_grad}

\begin{table*}[tb]
\centering
\resizebox{0.8\textwidth}{!}{
\begin{tabular}{ccC{1.5cm}C{1.5cm}C{1.5cm}C{1.5cm}|C{1.5cm}C{1.5cm}|C{1.6cm}}
\hline
 &  & C.man & house & Lena & boat & house.c & peppers& avg. \\ \hline
\multirow{2}{*}{Gaussian} & reg & 24.108 & 29.541 & 29.663 & 26.353 & 27.842 & 28.550 & 27.676 \\
 & Ours &\textBF{25.093}  &\textBF{30.745} & \textBF{30.705} &\textBF{27.436}& \textBF{29.021}& \textBF{28.827}&\textBF{28.638}\\ \hline
\multirow{2}{*}{Defocus} & reg & 23.841 & 29.053 & 29.164 & 25.874 & 27.488 & 28.210 &27.272 \\
 & Ours &\textBF{25.688}  &\textBF{30.473} &\textBF{30.355}  &\textBF{27.480} &\textBF{29.594}&\textBF{29.089}&\textBF{28.780}  \\\hline
\multirow{2}{*}{Motion blur} & reg & \color{red}{6.921} & \color{red}{6.142} & \color{red}{5.251} & \color{red}{6.268} & \color{red}{6.172} & \color{red}{5.697} &\color{red}{6.075}\\
 & Ours & \textBF{27.089} &\textBF{31.566}  & \textBF{31.801} & \textBF{28.435}&\textBF{30.007}&\textBF{29.661}&\textBF{29.760}
\end{tabular}
}
\caption{Kernel-known deconvolution \textit{PSNR} comparison between baseline (denoted by reg above) and ours.}
\label{tab:kn_psnr}
\end{table*}
\begin{table*}[tb]
\centering
\resizebox{0.8\textwidth}{!}{
\begin{tabular}{ccC{1.5cm}C{1.5cm}C{1.5cm}C{1.5cm}|C{1.5cm}C{1.5cm}|C{1.6cm}}
\hline
 &  & C.man & house & Lena & boat & house.c & peppers& avg. \\ \hline
\multirow{2}{*}{Gaussian} & reg & 19.553 & 14.214 & \textBF{29.798} & \textBF{26.323} & 14.662 & \textBF{24.790}&21.557 \\
 & Ours & \textBF{23.230} & \textBF{27.748} &26.094 & 24.977 & \textBF{27.122}&21.347&\textBF{25.086}\\ \hline
\multirow{2}{*}{Defocus} & reg & 18.845 & 13.519 & \textBF{27.435} & 24.035 & 13.849  & 24.782&20.411\\
 & Ours & \textBF{23.021} & \textBF{23.094} & 26.286 & \textBF{25.154}& \textBF{24.462} & \textBF{28.229}&\textBF{25.041}\\ \hline
\multirow{2}{*}{Motion blur} & reg & 16.835 & 12.865 & 25.304 & 22.625 & 15.295 & 22.207&19.189\\
 & Ours &\textBF{23.935}  &\textBF{24.382} & \textBF{26.156} & \textBF{25.039}& \textBF{22.862}& \textBF{26.152} & \textBF{24.754}
\end{tabular}
}
\caption{Blind deconvolution \textit{PSNR} comparison between baseline (denoted by reg above) and ours.}
\label{tab:blind_psnr}
\end{table*}

\begin{figure}[tb]
\centering
\includegraphics[width=.9\columnwidth]{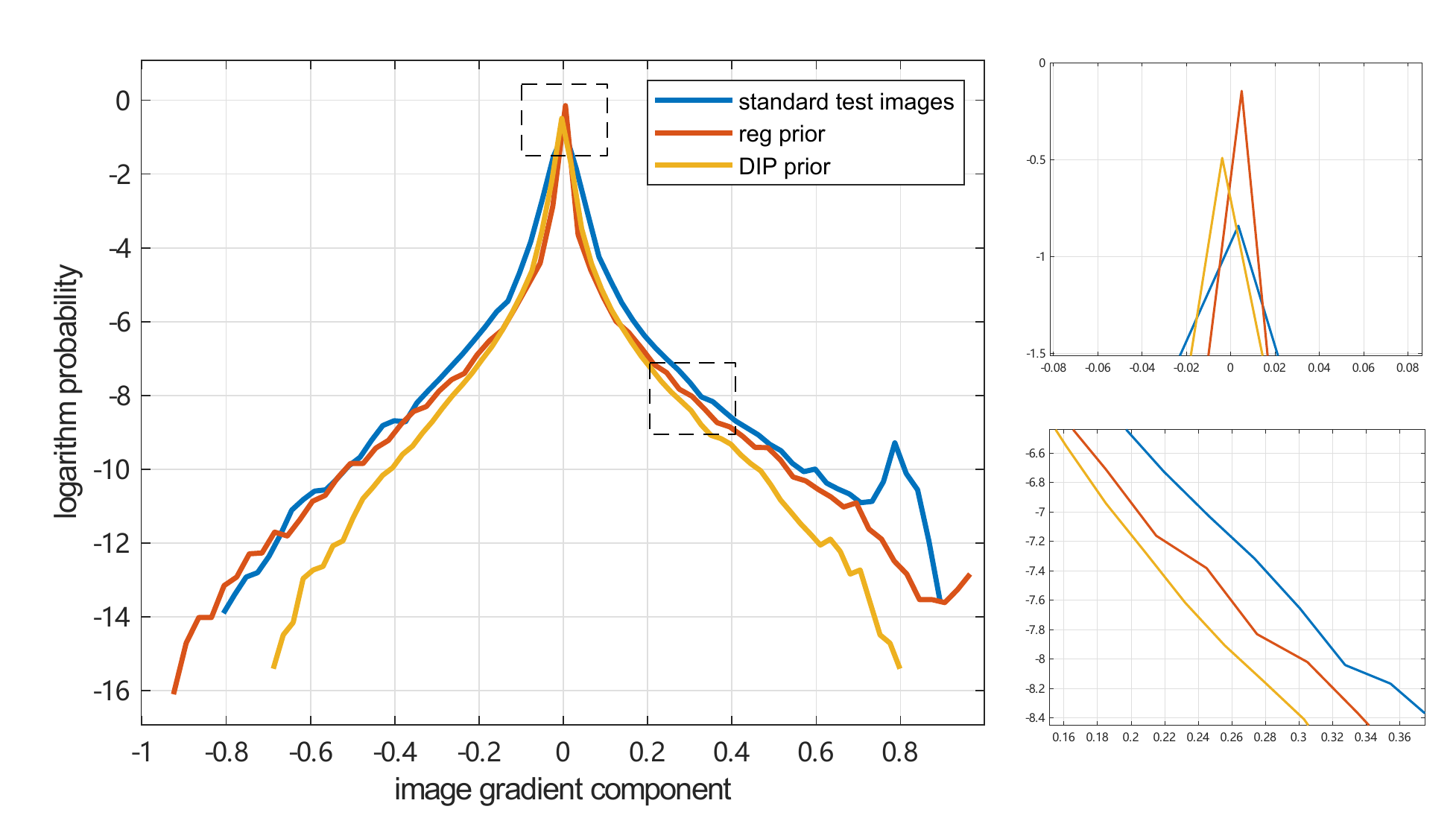}
\caption{Log-probability distributions of image gradients.}\label{fig:grad_dist}
\end{figure}
Previous image statistics studies \cite{WeissF07, Roth:CCVPR:05} have shown that natural image gradients follow heavy-tailed distributions, which provide a \textit{natural} prior for natural images. Starting from this, we consider evaluating the gradient distributions of our model-generated images with a ``standard'' distribution which can be assumed as the \textit{natural} prior.

With notations in \autoref{subsec:baseline}, the gradients of image $\mathbf{X}$ can be defined as matrices $\mathbf{X}\mathbf{D}_{1, n}^\top$ (horizontal) and $\mathbf{D}_{1, m}\mathbf{X}$ (vertical) \cite{di1986note}, where each element is a gradient value. In this experiment, we calculate the image gradient value distributions in $3$ image sets, standard test images, images by the baseline model and images by the DIKP model. The estimated probability distribution from frequency for each set is denoted by $\hat{\Pr}_{\rm{std}}\left(\cdot\right)$, $\hat{\Pr}_{\rm{reg}}\left(\cdot\right)$ and $\hat{\Pr}_{\rm{dikp}}\left(\cdot\right)$, where $\hat{\Pr}_{\rm{std}}\left(\cdot\right)$ is assumed to be the ``standard'' distribution. Therefore between the distributions by the $2$ model-generated image sets, the one with greater similarity to the ``standard'' distribution is more in line with the \textit{natural} prior.

Since the values of image gradients are continuous because of their double-precision floating-point data type, we split the range of gradient values $\left[-1, 1\right]$ into $64$ disjoint bins and count the number of gradient values that fall in each bin as the frequency. \autoref{fig:grad_dist} plots the logarithm probability distribution for each image set. Since the plot is in log scale, we can infer that all the three distributions have the heavy-tailed property, and their log-probability curves are similar in shape to each other. The peak close-up in the distribution shows a decreasing order of baseline-DIKP-standard in terms of log-probability, the gradient values in which lie around $0$. This shows that the density of the baseline and DIKP model where the gradient values are close to $0$ is larger than the standard images, and further speaking, the DIKP model performs closer to the standard than the baseline in this range. However, the close-up in the middle of peak and tail gives an order of standard-baseline-DIKP, which indicates the exact opposite to the above peak-range results. The results above are in expectation because the TV regularizer in the baseline tends to reduce image gradient values due to the property of TV norm \cite{chan2005recent} and thereby gives high frequency where gradients are close to $0$, and low frequency outside of peak range, which also illustrates DIKP's better performance in high frequency gradients.

Overall, the \textit{KL} divergence between gradient distributions of DIKP-generated images and standard test images is $\kl{\hat{\Pr}_{\rm{dikp}}}{\hat{\Pr}_{\rm{std}}} = 0.954$, while for the baseline, $\kl{\hat{\Pr}_{\rm{reg}}}{\hat{\Pr}_{\rm{std}}} = 1.260$. This indicates that DIKP have a greater similarity to the ``standard'' than the baseline in terms of gradient distribution. The result is foreseeable because although the baseline performs closer to the standard than DIKP in the middle range, DIKP perform closer to the standard in the peak with much higher frequency.% Therefore, DIP model is more capable of capturing natural image gradient prior compared to the baseline.

\subsection{Performance on deconvolution}\label{subsec:performance}
\begin{figure*}[tb]
\minipage{0.19\textwidth}
  \includegraphics[width=\linewidth]{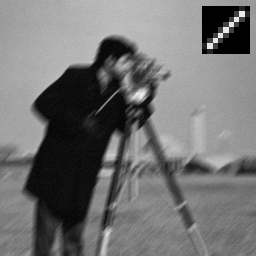}
  \subcaption{motion blurred}
\endminipage\hfill
\minipage{0.19\textwidth}
  \includegraphics[width=\linewidth]{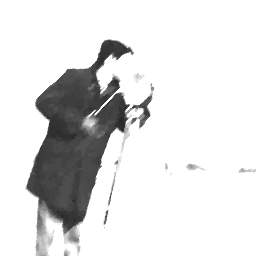}
  \subcaption{kernel-known baseline}\label{subfig:cmm_kn_baseline}
\endminipage\hfill
\minipage{0.19\textwidth}%
  \includegraphics[width=\linewidth]{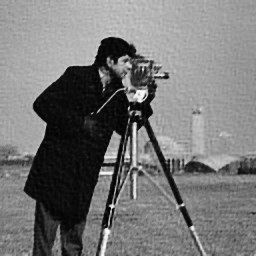}
  \subcaption{Ours (kernel-known)}\label{subfig:cmm_kn_DIP}
\endminipage\hfill
\minipage{0.19\textwidth}%
  \includegraphics[width=\linewidth]{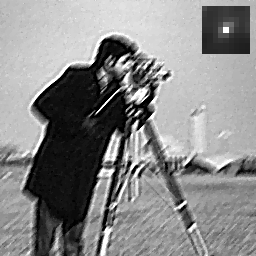}
  \subcaption{blind baseline}\label{subfig:cmm_blind_baseline}
\endminipage\hfill
\minipage{0.19\textwidth}%
  \includegraphics[width=\linewidth]{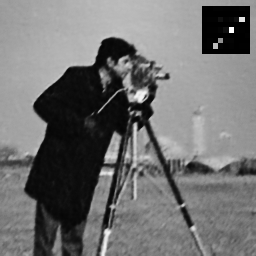}
  \subcaption{Ours (blind)}\label{subfig:cmm_blind_DIP}
\endminipage
\caption{Comparison on motion blurred \texttt{cameraman} between baseline and DIKP in both kernel-known and blind settings.}\label{fig:comp_cmm}
\end{figure*}
\begin{figure}[tb]
\minipage{0.27\columnwidth}
  \includegraphics[width=\linewidth]{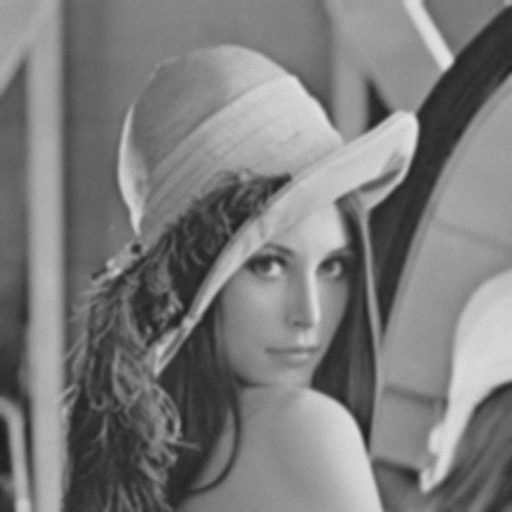}
  \subcaption{Gaussian}
\endminipage\hfill
\minipage{0.34\columnwidth}
  \includegraphics[width=\linewidth]{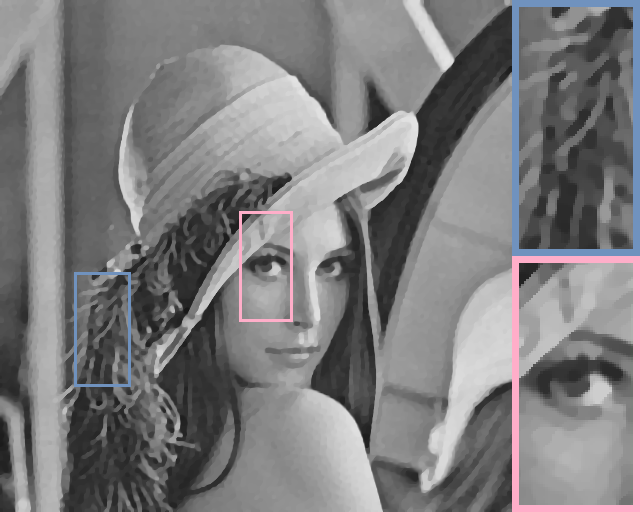}
  \subcaption{baseline (kk)}\label{subfig:lena_baseline_kk}
\endminipage\hfill
\minipage{0.34\columnwidth}%
  \includegraphics[width=\linewidth]{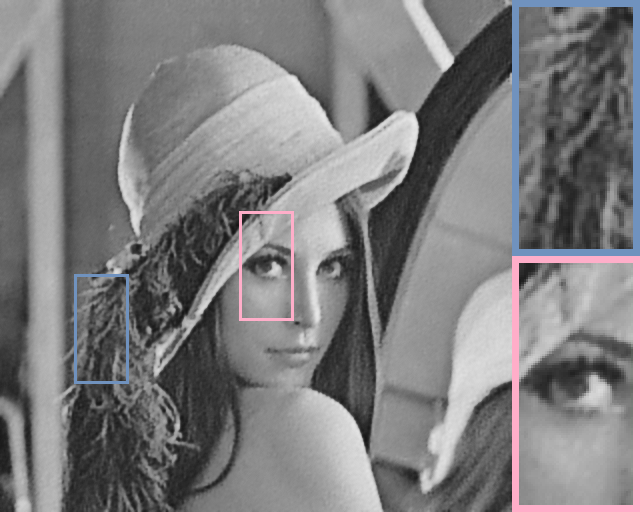}
  \subcaption{Ours (kk)}\label{subfig:lena_DIP_kk}
\endminipage
\newline\vskip 1mm
\minipage{0.27\columnwidth}
  \includegraphics[width=\linewidth]{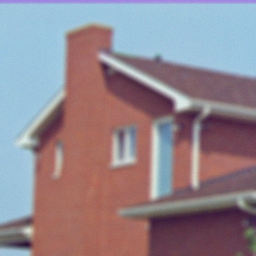}
  \subcaption{defocused}
\endminipage\hfill
\minipage{0.34\columnwidth}
  \includegraphics[width=\linewidth]{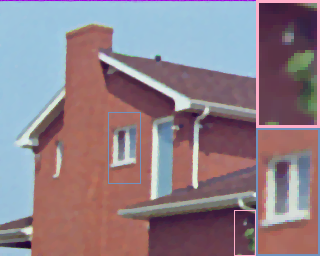}
  \subcaption{baseline (kk)}\label{subfig:house_baseline_kk}
\endminipage\hfill
\minipage{0.34\columnwidth}%
  \includegraphics[width=\linewidth]{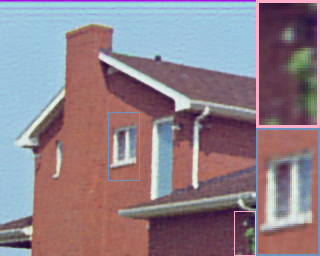}
  \subcaption{Ours (kk)}\label{subfig:house_DIP_kk}
\endminipage
\caption{Comparison on Gaussian degraded \texttt{Lena}, defocused \texttt{house.c} between baseline and deep image prior in kernel-known (abbriviated as kk above) setting.}\label{fig:comp_kk}
\end{figure}
We run our baselines and DIKP models on $18$ degraded images ($3$ degradation kernels on $6$ standard test images) in both kernel-known and blind settings. Then we compute the \textit{PSNR} between generated results and original standard test images, and visualize some of the results for quantitative and qualitative comparison respectively.

Shown in \autoref{tab:kn_psnr} and \autoref{tab:blind_psnr} are \textit{PSNR} comparisons between baseline and deep priors 
for kernel-known and blind deconvolution respectively. Overall, our DIKP deconvolution models always perform better than baseline models in terms of average \textit{PSNR} on different degradation kernels. In kernel-known setting, DIKP even give a larger \textit{PSNR} value on every single degraded image. Particularly, when the kernel type is set as motion blur, the baseline gives unexpectedly bad results as shown by the \textit{PSNR} values marked in red in \autoref{tab:kn_psnr}. We suspect this is because TV regularizer overfits the gradient prior on the motion deblur, so that the non-edge regions of the image tend to be in the same pixel value (see \autoref{subfig:cmm_kn_baseline}). When the kernel is set to Gaussian or defocus, the performance is improved by around $1.2\pm0.3$ in terms of \textit{PSNR} as we expect. In blind setting, DIKP improve the \textit{PSNR} performance by around $5.0\pm0.5$, which is significantly beyond the performance of the baseline. However, baseline gives higher \textit{PSNR} values than the deep image prior for a few pictures and kernel types, such as \texttt{Lena} degraded by Gaussian or defocus. A possible reason is that the gradient values in \texttt{Lena} are relatively small, so that TV regularization gives better results on this specific image.

\autoref{fig:comp_kk} visualizes the comparison between images restored from Gaussian degraded \texttt{Lena} and defocused \texttt{house.c} in kernel-known setting. From the pictures and their close-ups, we see that DIKP perform better in detail recovery. For example, the hair in \autoref{subfig:lena_baseline_kk} has only a clear outline, while the details shown in \autoref{subfig:lena_DIP_kk} are more abundant as well as the trees shown in \autoref{subfig:house_DIP_kk} compared with \autoref{subfig:house_baseline_kk}. One possible explanation is that TV regularizer over-optimizes the sharpness of images, resulting in good performance only in outlines but not in detail.

In spite of the two kernels above, DIKP achieve remarkable results especially in motion blur deconvolution. \autoref{fig:comp_cmm} visualizes the comparison between images restored from motion blurred \texttt{C.man} in both settings. As mentioned previously, kernel-known baseline gives an unsatisfactory result (\autoref{subfig:cmm_kn_baseline}), where only the basic outline of the cameraman can be observed, and all other details inside the image are lost, while kernel-known DIKP restore the image almost perfectly as shown in \autoref{subfig:cmm_kn_DIP}. For blind motion deblurring on \texttt{C.man}, The result (\autoref{subfig:cmm_blind_baseline}) given by baseline still has motion blur, and the shape of its kernel is completely different from motion blur, while DIKP remove motion blur efficiently and the shape of its kernel is much closer to motion blur than the baseline (see \autoref{subfig:cmm_blind_DIP}), which also verifies ConvNet's expressiveness on degradation kernels.

%------------------------------------------------------------------------

\section{Conclusions}
\label{sec:concl}
We investigate deep ConvNet's expressiveness on the prior information of natural images and degradation kernels in DIKP image deconvolution, and present its performance in both kernel-known and blind settings. More importantly, we propose DIKP-based energy minimization pipelines for image deconvolution in the two settings, and achieve performance which is far beyond our baselines \cite{beck2009fast, wang2017iterative}. Our motivation is to adopt DIKP with more complex structures to express image prior information based on the idea of traditional learning-free optimization methods, and at the same time to improve image deconvolution performance by traditional learning-free methods. Through the first two experiments, we prove that the ConvNet structures of DIKP capture strong prior information on natural images in terms of generation types and gradient distributions. In the final experiment, we show the significant improvement by DIKP models compared with the baselines in terms of both \textit{PSNR} values and visual effects, especially for motion-blurred images. However, we verify DIKP's expressiveness on degradation kernels only by an adjusted \textit{hourglass} structure. It is hard to associate kernel features and deep neural structures intuitively. Therefore, future endeavours in this topic should focus on the structures of DIKP generating kernels, trying other hyperparameters on \textit{hourglass}, or other ConvNet structures, e.g. \textit{texture nets} \cite{ulyanov2016texture}. Besides, as applied in \cite{shan2008high}, the formulation of energy functions may be adjusted with gradient terms to become more suitable for this task.

%------------------------------------------------------------------------
\medskip
\noindent \textbf{Acknowledgements.} We thank Yusheng Tian for helpful changes and Prof. Steve Renals for organizing this project.

{\small
\bibliographystyle{ieee}
\bibliography{egbib}
}

\end{document}